\documentclass[conference]{IEEEtran}
\usepackage{stfloats}
\usepackage{cite}
\usepackage{amsmath,amssymb,amsfonts}
\usepackage{algorithmic}
\usepackage{graphicx}
\usepackage{textcomp}
\usepackage{xcolor}
\usepackage{hyperref}
\hypersetup{hidelinks,
	colorlinks=true,
	allcolors=black,
	pdfstartview=Fit,
	breaklinks=true}
\ifCLASSINFOpdf
\else
\fi
\begin{document}
\title{CvT-ASSD: Convolutional vision-Transformer Based Attentive Single Shot MultiBox Detector}

\author{\IEEEauthorblockN{1\textsuperscript{st} Weiqiang Jin}
\IEEEauthorblockA{\textit{School of Computer Engineering} \\
\textit{and Science, University} \\
\textit{Shanghai, China} \\
Postal Code: 200444\\
Email: weiqiangjin@shu.edu.cn}
\and
\IEEEauthorblockN{2\textsuperscript{nd} Hang Yu}
\IEEEauthorblockA{\textit{School of Computer Engineering} \\
\textit{and Science, University} \\
\textit{Shanghai, China} \\
Postal Code: 200444\\
Email: yuhang@shu.edu.cn}
\and
\IEEEauthorblockN{3\textsuperscript{rd} Xiangfeng Luo}
\IEEEauthorblockA{\textit{School of Computer Engineering} \\
\textit{and Science, University} \\
\textit{Shanghai, China} \\
Postal Code: 200444\\
Email: luoxf@shu.edu.cn}}


%Telephone: (800) 555--1212, Fax: (888) 555--1212}


\maketitle

\begin{abstract}
Due to the success of Bidirectional Encoder Representations from Transformers (BERT) in natural language process (NLP), the multi-head attention transformer has been more and more prevalent in computer-vision researches (CV). However, it still remains a challenge for researchers to put forward complex tasks such as vision detection and semantic segmentation. Although multiple Transformer-Based architectures like DETR and ViT-FRCNN have been proposed to complete object detection task, they inevitably decreases discrimination accuracy and brings down computational efficiency caused by the enormous learning parameters and heavy computational complexity incurred by the traditional self-attention operation. In order to alleviate these issues, we present a novel object detection architecture, named Convolutional vision Transformer-Based Attentive Single Shot MultiBox Detector (CvT-ASSD), that built on the top of Convolutional vision Transormer (CvT) with the efficient Attentive Single Shot MultiBox Detector (ASSD).  We provide comprehensive empirical evidence showing that our model CvT-ASSD can leads to good system efficiency and performance while being pretrained on large-scale detection datasets such as PASCAL VOC and MS COCO. Code has been released on public github repository at \href{https://github.com/albert-jin/CvT-ASSD}{https://github.com/albert-jin/CvT-ASSD}.
\end{abstract}



% \fi
\IEEEpeerreviewmaketitle

\section{Introduction}
Real-time Object Detection task is challenging yet essential in computer vision researches.The target of object detection is to determine a set of bounding boxes and corresponding category labels for each object of interest presented in pictures.Thanks to many advantages of convolution such as local receptive, spatial subsampling and shared weights which could preserve rich semantic information during the deep-learning network forward flow operations, convolution-based architectures remain dominant\cite{leCunHandwrittenZipCode} for decades. 

 In recent years after BERT \cite{vaswani2017attentionallyouneed} provided by Google, Transformer-based architecture has become the leading technology in many NLP tasks due to its powerful language understanding performance borrowed by multiheads self-attention module. Inspired by success in NLP, much of the recent progress made in object detection research can be credited to applying transformer model to translate vision representations learned on massive object detection datasets. ViT \cite{dosovitskiy2021ViT}, the first attempt of  Self-Attention-based visual representation learning, which explicitly model all pairwise interactions between elements in a sequential embedding vector, demonstrates that transformer-based architectures can improve both image classification performance and efficiency if pretrained on large-scale image datasets such as JFT-300M \cite{sun2017jft300m} and IG-940M.  \cite{mahajan2018ig940m}.

A wide range of object-detection approaches like vision-transformer based Faster-RCNN model (ViT-FRCNN) \cite{joshbeal2020vitfrcnn} and end-to-end object detector with Adaptive Clustering Transformer (ACT) \cite{minghang2020act}, which built on vision transformer that generating computer analytic semantic signals through streamline the training pipeline by viewing object detection as a direct anchors with labels prediction problem. This end-to-end philosophy has led to significant advantages in  complex structured vision tasks such as image retrieval,image segmentation and environment dense prediction. 

Despite the success of vision transformers at large scale, they both are vulnerable to low efficiency brought by huge training parameters inside transformer modules and poor model recognition performance when trained on smaller amount of data. Meanwhile, vision transformer suffers severely from the heavy computational complexity due to high-resolution image inputs in a few of downstream vision tasks. Given an H × W resolution picture, the learning parameter complexity for each multi-head attention module is $\mathcal{O}(H^2W^2d)$. In recent, many researchers focused on such challenges and provided many resolutions like the spatially separable self-attention (SSSA) \cite{xiangxiang2021twins}.

As far as we know, images have a strong 2D local structure: spatially neighboring pixels are usually highly correlated. Human object recognition ability relies heavily on the spatial characteristics of an object, and so do computer recognition. However, ViT \cite{dosovitskiy2021ViT} lacks certain desirable properties inherently built into CNN architecture. The CNN-based architectures \cite{ren2016fasterRCNN,girshick2014fastrcnn,liuwei2016ssd} uniquely suited to solve vision tasks because of their strong capability capturing local structure by using shared weights, spatial subsampling and local receptive fields. The pioneering work of ViT on image classification are encouraging, but its architecture is unsuitable for use as a general-purpose backbone network on dense vision tasks due to its quadratic increase in complexity with high resolution image as input. Furthermore, despite the superiority of vision transformers when they were pretrained in large scale dataset, those comprehensive performances including accuracy and efficiency are still worse than other similar Convolution-based network architectures like VGG-Net \cite{simonyan2014deep}, Faster-RCNN \cite{ren2016fasterRCNN} and ResNet \cite{KaimingResNet}.

\begin{figure*}[ht]
\centerline{\includegraphics[width=1.0\linewidth]{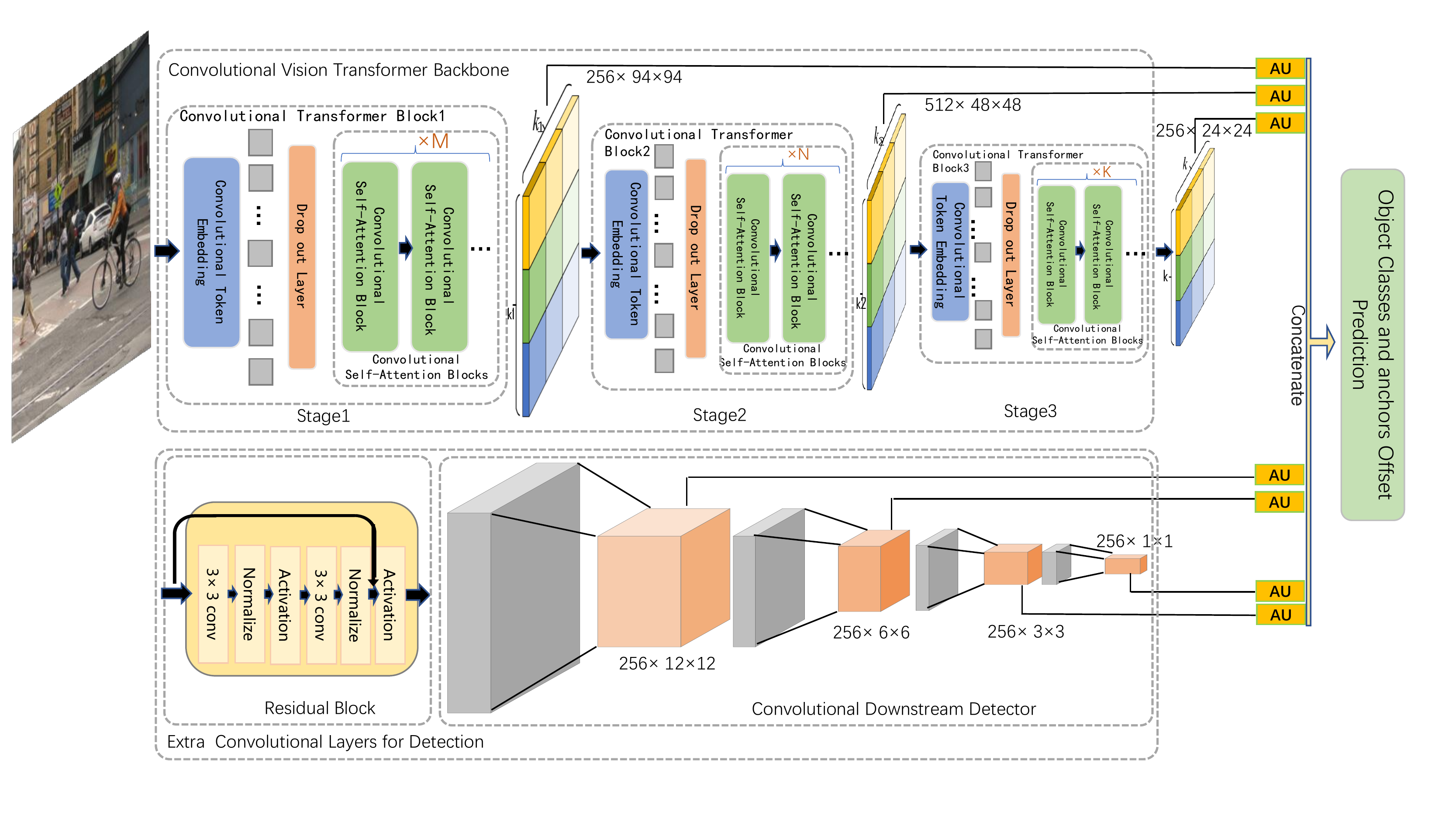}}
\caption{The overall pipeline of our proposed CvT-ASSD architecture. We feed an image vector to the Convolutional vision Transformer (CvT) backbone network which composed of hierarchical multi-stage Convolutional Transformer Block (CTB). It generates feature map represented for shallow semantics in each of three stages, and then makes the resulting vector flow through intermediate Residual Block (RB) between the transformer encoder and the detection module. Finally, we extract the abstract semantics through the feature maps generated by the detection module which composed of several pyramid convolutional blocks. All of the resulting feature map vectors, which maintain rich visual semantics, flow into Attentive Unit (AU) to guide the detection with refined information and followed the standard prediction step as naive SSD \cite{liuwei2016ssd}. Details of CTB and AU modules are shown in Figure~\ref{ConvTransformerBlock} and Figure~\ref{AttentionUnit}, respectively.}
\label{overall_architecture}
\end{figure*}

To overcome these issues, in this paper, we propose a novel Transformer-based approach for object detection that introduces convolutions to vision transformer backbone network and add self-attention mechanism into downstream detector SSD, called Convolutional vision Transformer-based Attentive Single Shot MultiBox Detector (CvT-ASSD). The overall model architecture of our proposed CvT-ASSD is shown in Figure~\ref{overall_architecture}. Convolutions were introduced into the original Vision Transformer architecture (ViT) \cite{dosovitskiy2021ViT} to merge the benefits of transformers with the benefits of CNNs for image object detection task. It replaces the traditional self-attention module on the original ViT \cite{dosovitskiy2021ViT} framework with a new self-attention module that calculating area attention weights by convolutional query·key·value operations. In addition, we also adapt the self-attention mechanism to our downstream detector module, named Attention Unit, which helps to highlight useful regions on the feature maps while suppressing the irrelevant information, thereby providing reliable guidance for object detection. We compare our method with state-of-the-art object detection methods including VGG-SSD, DETR and ViT-FRCNN evaluating on the most popular object detection datasets, PASCAL-VOC \cite{Everingham15voc} and COCO \cite{lin2015mscoco}. Experiments show that our model achieves comparable mean-average-Precision (maP) performances with fewer parameters and FLOPs. Model implementation details will introduce in Chapter \ref{section3}, Our Model.

Our main contributions are summarised as follows:

1. We propose a novel unified object detection architecture, CvT-ASSD, which modifies transformer backbone module by adding the convolutional token embedding and convolutional projection into transformer encoder block, along with the multi-stage design of the network by convolutions, making our model achieve superior performance while maintaining certain computational efficiency.

2. We apply an Residual Block (RB) between the transformer encoder and the downstream modules which can help avoid the degradation problem. Ablation experiments show that this module can lead to a significant average precision boost for the whole architecture performance.

3. We introduce the self-attention mechanism to downstream detection module termed ASSD following the human vision mechanism and facilitates the object feature learning. It effectively utilizes a fast and light-weight attention unit to help discover feature dependencies and focus the model on useful and relevant regions.

4. Specifically, we evaluate our CvT-ASSD and its variants on the most popular detection challenge, COCO and VOC. Extensive experiments show that our proposed architecture performs favorably against other state-of-the-art vision transformer with similar or even reduced computational complexity.

Furthermore, we hope that CvT-ASSD can drive the commonly applied paradigm of large scale pretraining and rapid fune-tuning to specific tasks deeper and encourage Transformer-based unified modeling in the computer vision community.

\section{Related Work}

\subsection{Traditional object detector based on convolution backbone network}

Faster-RCNN \cite{ren2016fasterRCNN}, one comparable model in the series of region-base CNN, which introduce novel Region Proposal Network (PRNs) that share convolutional layers with modern object detection networks SPP-net \cite{heKaiming2014SPPnet} and Fast-RCNN \cite{girshick2014fastrcnn}, improves region proposal quality and thus overall object detection accuracy. Due to the computational cost-free region proposal step inside Faster-RCNN architecture, the method enables a unified, deep-learning-based object detection system to run at near real-time frame rates. Although Faster-RCNN achieves performance in a competitive rate than other traditional methods, there is still room for improvement in accuracy caused by the unnatural design of increasing translation variance. Meanwhile, the deep convolutional backbone network like Res-Net inside the architecture brings a large scale of calculating parameters which would hinder the speed in both training and inference.

To address the dilemma above in Faster-RCNN \cite{girshick2014fastrcnn}, a region-based, fully convolutional network (R-FCN) \cite{dai2016rfcn} presented by Microsoft Research team, which applies a costly per-region subnetwork hundreds of times in contrast to previous region-based detector such as Fast/Faster R-CNN \cite{girshick2014fastrcnn,ren2016fasterRCNN}, achieved at test-time speed of 170ms per image, 2.5-20× faster than Faster R-CNN counterpart. The R-FCN architecture is designed to classify the regions of Interests (RoIs) into object categories and background. RoIs is proposed by Region proposal network (RPN) through the predefined score maps.
% In the original paper, the backbone feature extractor of R-FCN incarnation is ResNet-101, and also other convolutional networks like VGG-Net \cite{simonyan2014deep} and Le-Net \cite{yannlecun1998lenet} are applicable in their experiments. 
The experiments on the R-FCN paper empirically justifies the importance of respecting spatial information by inserting RoIs pooling between layers for the Faster R-CNN system.

Single Shot MultiBox Detector (SSD) \cite{liuwei2016ssd}, put forward by Google Inc, is significantly more accurate and faster than the previous state-of-the-art object detectors like YOLOs, in fact as accurate as slower techniques that perform explicit region proposals and pooling (including Faster RCNN). The SSD approach is based on a feed-forward convolutionval network that produces a fixed-size collection of anchor boxes and scores for the presence of object instances in those anchor boxes, followed by a non-maximum suppression (NMS) to filter out the final predict results. These creative model features lead to easy end-to-end training and high testing accuracy meantime, further improving the speed vs accuracy trade-off.

\subsection{Naive Vision Transformer Based object detector}

 DEtection TRansoformer (DETR) \cite{FacebookAI2020detr}, a new method still built in the top of ViT \cite{dosovitskiy2021ViT}, which views object detection as a direct set prediction problem, demonstrates accuracy and run-time performance on par with the well-established and highly optimized Faster R-CNN baseline on the challenging COCO-2014 \cite{lin2015mscoco} and PASCAL VOC2007$\&$2012 \cite{Everingham15voc} object detection datasets \cite{Everingham15voc}. Unlike many other modern detectors, this model is conceptually simple and does not require a specialized library. A notable property in this approach is that it does not need to use non-maximum suppression (NMS) as a post-processing step, as its decoder architecture learns to self-suppress duplicate bounding box predictions. By the way, there are some shortcomings in this approach: 1) slow speed of training convergence than typical detectors and 2) limited feature spatial resolution when transformer processes image data. These two drawbacks mainly stem from prohibitive complexities in processing high-resolution feature maps.

The efficient object detection architecture codenamed ViT-FRCNN \cite{joshbeal2020vitfrcnn} that using original vision-transformer (ViT) \cite{dosovitskiy2021ViT} backbone to retain sufficient spatial information, which trained end-to-end with a set loss function which performs bipartite matching between predicted and ground-truth objects, finally achieved high accuracy, large pretraining capacity and fast superior fine-tuning performance. 
% Although it can be viewed as an important stepping stone toward a vision-transformer solution of complex vision task, it could decrease the model convergence speed due to the unshared per-RoI computation and independent region-wise based detection pattern.

Meanwhile, more and more powerful variants of vision Transformer-based architectures \cite{ictai2020idnet,liu2021swin,xiangxiang2021twins,ictai2020llsa,yuhang_svm,yuhang_csvm} are presented for image-level classification and a few downstream vision tasks, bringing continuous improvement in state-of-the-art object detection performance.

\section{Our Method: CvT-ASSD}
\label{section3}

In this section, we first revisit the overall pipeline of our proposed Transformer-based one-stage detector: CvT-ASSD in Section \ref{section3.1}. Implementation details and hyperparameter settings are presented in this section. Then, in Section \ref{section3.2} we introduce the details of the novel Convolutional vision Transformer which include Convolutional Token Embedding module and Convolutional Self-Attention module. Finally, In Section \ref{section3.3}, we provide a comprehensive analysis about the superiority that applies Residual Block (RB) and attention unit (AU) before and after the detection module, respectively. The full model structure is built on DeepLearning framework PyTorch v1.9.0 and is open source at: \href{https://github.com/albert-jin/CvT-ASSD}{https://github.com/albert-jin/CvT-ASSD}.

\subsection{Main Structure of CvT-ASSD}

The overall pipeline of CvT-Based Attentive Single Shot MultiBox Detector (CvT-ASSD) is illustrated in Figure~\ref{overall_architecture}. Our introduced CvT-ASSD is a competitive object detection solution which utilizes convolution in transformer MHSA part and attention operation in downstream detection step. The model architecture can be split into several relatively independent modules in turn: {\sffamily Convolutional Vision Transformer-\textgreater Residual Block-\textgreater Convolutional Downstream Detector-\textgreater Attention Unit-\textgreater SSD Standard Optimizer}. Next paragraphs we will discuss these modules in detail.

In the Convolutional Vision Transformer feature extractor, we introduce two convolution-based operations into each blocks. We term the two calculation modules as Convolutional Token Embedding (CTE) and Convolutional Self-Attention (CSA), respectively.

CTE is implemented as a 2D convolution operation with overlapping patches of which convolution kernel size is 7×7 and stride is 3×3. This allows each stage of the vision transformer backbone to progressively reduce the number of token such as image resolution and feature channels.

To obtain the ability to capture local spatial relationships throughout the ViT work \cite{dosovitskiy2021ViT} like CNNs, we changed previous vision Transformer modules by replacing the position-wise linear projection with our convolutional projection CSA. CSA is implemented using a depth-wise separable convolution layer to replace the original position-wise linear projection for Multi-Head Self-Attention (MHSA) in the ViT \cite{dosovitskiy2021ViT} work. Furthermore, these two built-in properties give us the ability to capture local spatial relationships and global semantic context throughout the network which allows us to discard the position embedding from the transformer, so we drop the positional embedding for tokens without hurting performance. The resulting new Transformer Block with the convolutional Projection layer is a generalization of the original ViT \cite{dosovitskiy2021ViT} design.

In the extra layers of downstream detector between the transformer encoder and the detection module, we add an intermediate Residual Block (RB) which put forward in work\cite{KaimingResNet} by KaimingHe et al. Residual block can help avoid the degradation problem: Typically, with the network depth increasing, accuracy gets saturated (which might be unsurprising) and then degrades rapidly. And in this work, we investigate the impact of the added residual block and find that introducing the module leading to a significant average precision boost.

SSD \cite{liuwei2016ssd} performs the detection on multi-scale feature maps to handle various object size effectively, so in our downstream detection module, the pyramid convolutional blocks for detecting objects follow similar with the design of original SSD. The differences of feature map size and channel number between our detection module and original SSD are listed in Table~\ref{table_diff_convs}. Then, We apply convolution layers of 3×3×channels kernel in each feature layers to produce either a score for a object category or a shape offset relative to the default box coordinates.

Inspired by the superiority of self-attention mechanism in Transformer, we construct a small network, namely Attention Unit (AU), and embed it into the last layer of the downstream detection module to improve the detection accuracy. The AU module helps capture the long-range dependencies among all feature pixels within the feature map itself for more effective object detection.

In the end, we concatenate all of the resulting feature token into a one-dimension vector for location and object label prediction. The overall objective loss function is a weighted sum of the object label confidence loss (e.g. Softmax CrossEntropy Loss) and the localization loss (e.g. Smooth L1 Loss) followed by original SSD \cite{liuwei2016ssd}. The loss function and back propagation are applied end-to-end.
We use hard negative mining to solve the positive-negative box class imbalance problem and training process also involves multi-scales detection results, and data augmentation strategies as in original SSD.
\label{section3.1}

\setlength{\tabcolsep}{1.0mm}
\begin{table}
\centering
\caption{Differences of detected feature scale between original SSD and our CvT-ASSD. N/A means that original SSD gets detection results through only six convolutional layers. Data format ($A^2*B$) denotes the scale of Width*Height*Channel of each feature map.}
\label{table_diff_convs}
\begin{tabular}{|l|c|c|}
\hline
\textbf{Conv Layer}  &  \textbf{Origin SSD} & \textbf{CvT-ASSD}\\ \hline
Conv\_1 & $38^2$*512 & $94^2$*192 \\
Conv\_2 & $19^2$*1024 & $48^2$*768 \\
Conv\_3 & $10^2$*256 & $24^2$*1024 \\
Conv\_4 & $5^2$*256 & $12^2$*256 \\
Conv\_5 & $3^2$*256 & $6^2$*256 \\
Conv\_6 & $1^2$*256 & $3^2$*256 \\
Conv\_7 & N/A & $1^2$*256 \\
\hline
\end{tabular}
\end{table}

\subsection{Convolutional vision transformer}
\label{section3.2}

\begin{figure*}[ht]
\centering
\includegraphics[width=0.9\linewidth]{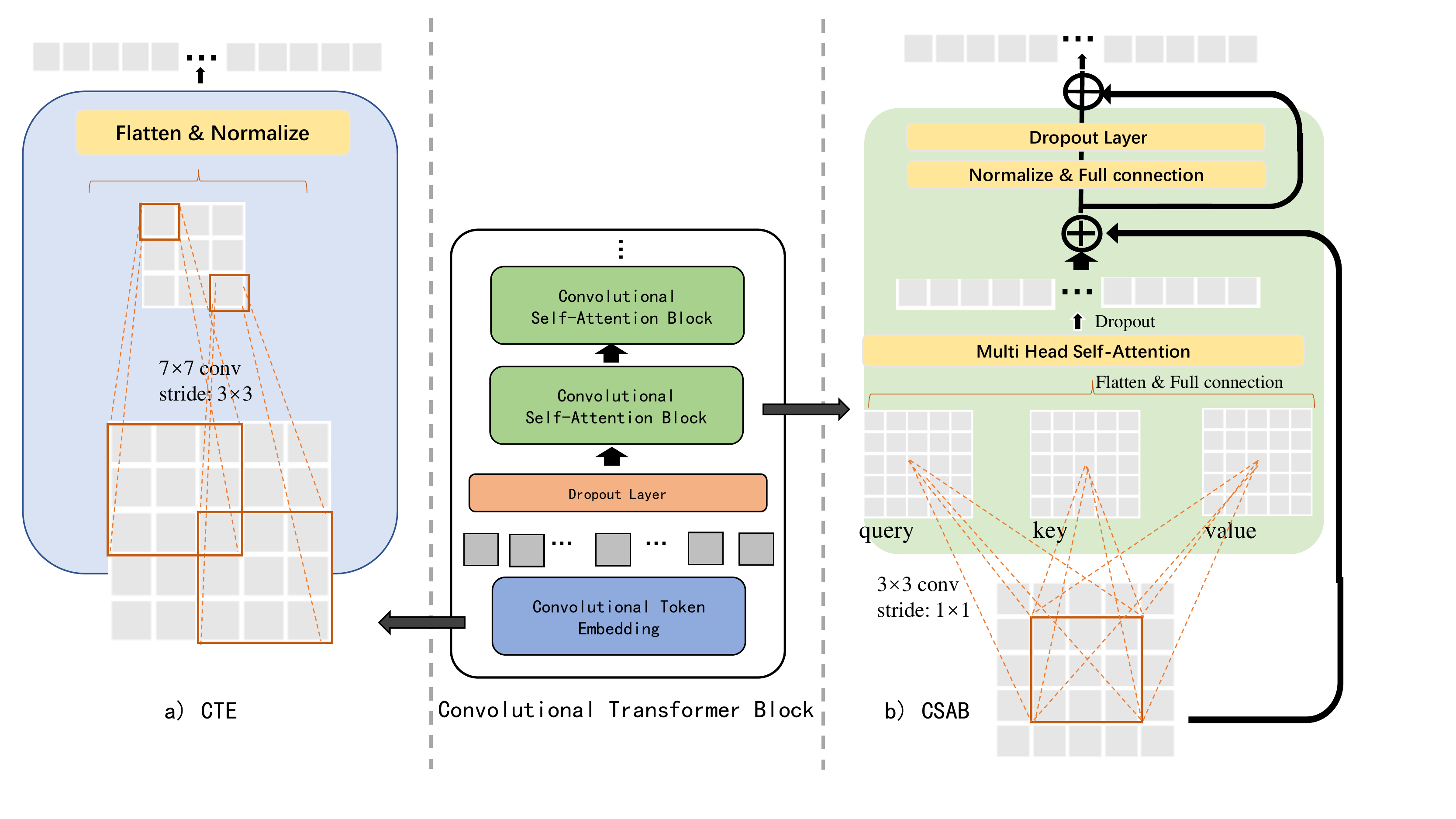}
\caption{Key idea of introducing convolutions into transformer. In this illustration, details of Convolutional Token Embedding is depicted in part $a$ (left), details of Convolutional Self-Attention Block is depicted in part $b$ (right), respectively.}
\label{ConvTransformerBlock}
\end{figure*}

Our CvT receives a 2D image vector as input. It consists of three module stages, termed Convolutional Transformer Block (CTB). We instantiate model with different parameters and FLOPs by varying  the hidden feature dimension and the number of Convolutional Self-Attention Block. In each transformer block, we progressively decrease the feature map size, while simultaneously increasing the feature map dimension. Furthermore, different from other prior Transformer-based architectures \cite{joshbeal2020vitfrcnn,xiangxiang2021twins,FacebookAI2020detr}, we discard the ad-hod position embedding to the tokens. Figure~\ref{ConvTransformerBlock} shows the internal structure details of transformer block. 

\subsubsection{Conv-Token Embedding in Transformer}
The Convolutional Token Embedding (CTE) layer allows us to regulate the feature map dimension and size at each stage by varying parameters of the convolution operation. This helps the model capture the increasing complex visual patterns over increasing larger spatial footprints, similar to CNN based feature extractor. Formally, given a 2D feature map $x_{i-1} \in \mathbb{R}^{H_{i-1} \times W_{i-1} \times C_{i-1}}$ generated from previous step as the input to CTE layer, we learn a mapping function $f(\cdot)$ that maps $x_{i-1}$ into a new tokens $f(x_{i-1})$ with a channel size $C_{i}$,where $f(\cdot)$ is 2D convolution operation of kernel size $k$ (equal 7×7), stride $s$ (equal 3×3) and padding $p$ (equals 1×1) (to handle the boundary conditions). The new feature map $f\left(x_{i-1}\right) \in \mathbb{R}^{H_{i} \times W_{i} \times C_{i}}$ has sizes $H_i, W_i$ :
\begin{equation}
H_{i}=\left\lfloor\frac{H_{i-1}+2 p-k}{s}+1\right\rfloor, W_{i}=\left\lfloor\frac{W_{i-1}+2 p-k}{s}+1 \right\rfloor.
\end{equation}
$f(x_{i-1})$ is then flatten into size $H_i W_i×C_i$ and normalized by layer normalization for input into the subsequent Convolutional Self-Attention Block (CSAB). The structure of CTE is depicted in Figure~\ref{ConvTransformerBlock} (a).

\subsubsection{Conv-Self-Attention in Transformer}
The embedded module Convolutional Self-Attention Block (CSAB) aims to model local spatial contexts, from low-level edges to higher order semantic primitives, over a multi-stage hierarchy approach, similar to CNNs. Standard $q k v$ self-attention is a popular building block for neural architectures. For each element in an input sequence $z \in \mathbb{R}^{N×D}$, we compute a weighted sum over all values $v$ in the sequence. The attention weights $A_i j$ are based on the pairwise similarity between two elements of the sequence and their respective query $q^{i}$ and key $k^{j}$ representations, Self-Attention operation $f_{SA}(\cdot)$ uses the following formula :
\begin{equation}
[\mathbf{q}, \mathbf{k}, \mathbf{v}]=\mathbf{z} \mathbf{U}_{q k v} \quad \mathbf{U}_{q k v} \in \mathbb{R}^{D \times 3 D_{h}}
\end{equation}

\begin{equation}
f_{SA}(\cdot)=\mathbf{v} \cdot \operatorname{softmax}\left(\mathbf{q k}^{\top} / \sqrt{D_{h}}\right) \quad f_{SA}(\cdot) \in \mathbb{R}^{N \times N}
\end{equation}

Multi Head Self-Attention (MHSA) is an extension of SA in which we run $k$ self-attention operations called "multihead" in parallel, and project their concatenated output. To keep compute parameters constant when changing $k$, $D_h$ is typically set to $D/k$.
\begin{equation}
{f}_{MSA}(z)=\mathbf{U}_{m s a} \left[\mathrm{f}_{SA1}(z) ; \mathrm{f}_{SA2}(z) ; \cdots ; \mathrm{f}_{SA3}(z)\right]
% \quad \mathbf{U}_{m s a} \in \mathbb{R}^{k D_{h} \times D}
\end{equation}

Different from the standard MHSA, this work replaces the original Position-wise Linear Projection Mechanism with our depth-wise separable convolutional Self-Attention Block (CSAB) module, into the Transformer architecture. The convolutional projection of CSAB is depicted in Figure~\ref{ConvTransformerBlock} (b).

\subsection{Downstream Detection Modules}
\label{section3.3}
The downstream detector includes several modules respectively are Residual Block, Convolutional Downstream Detector, Attention Unit, Standard MultiBoxLoss Optimizer. Except for the introduced module Residual Block and Attention Unit, others are similar to original SSD \cite{liuwei2016ssd}.

\subsubsection{Residual Block Layer}
Inspired by the philosophy of ResNet \cite{KaimingResNet}: If identity mappings are optimal, the solvers may simply drive the weights of the multiple nonlinear layers toward zero to approach identity mappings, we adopt residual learning to our model between the transformer encoder and the detection module. The extra shortcut connections introduces neither extra parameter nor computation complexity. The extra layer is attractive and efficient in our comparisons between plain and residual networks. 

\begin{figure}[h]
\centering
\includegraphics[width=1.0\linewidth]{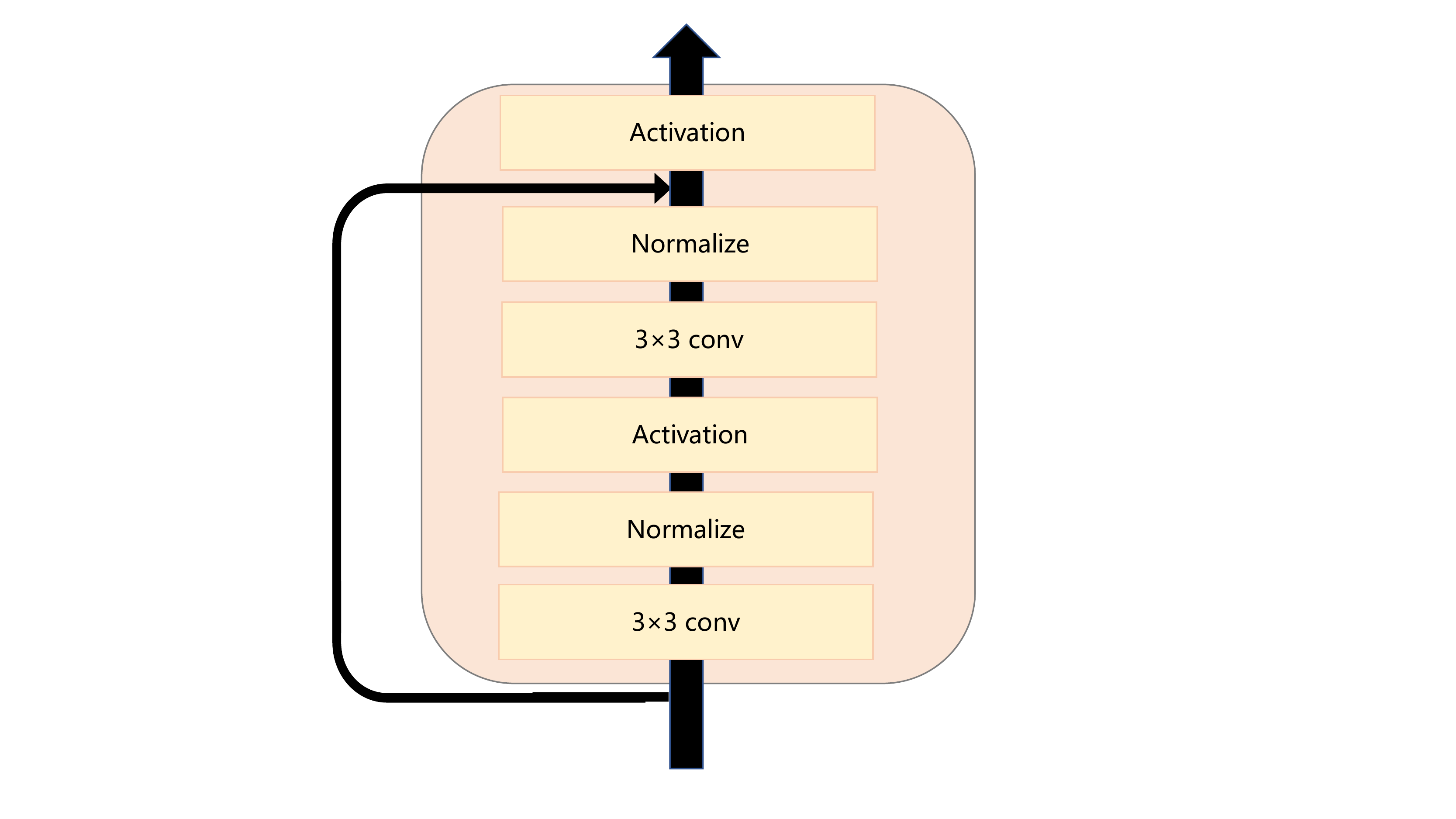}
\caption{In this figure, we illustrate the proposed Residual Block structure for solving the degradation problem. Inside the flow of residual mapping, we apply two 3×3 convolutions and two normalization layers which alternated with each other.}
\label{ResidualBlock}
\end{figure}

Let us consider $H(x)$ as an underlying mapping to be learn by a few neural network layers, with $x$ denoting the inputs to the first of these layers. Rather than approximating $H(x)$, we explicitly let these layers approximate a residual function $\mathcal{F}(\mathbf{x}):=\mathcal{H}(\mathbf{x})-\mathbf{x}$. So the original function will becomes $\mathcal{F}(\mathbf{x}) - \mathbf{x}$. Formally, in this work we define the residual mapping as:
\begin{equation}
\mathbf{y}=\mathcal{F}\left(\mathbf{x},\left\{W_{i}\right\}\right)+\mathbf{x}
\end{equation}
Here $\mathbf{y}$ and $\mathbf{x}$ are output vectors and input vectors, respectively. The functional mapping $\mathcal{F}\left(\mathbf{x},\left\{W_{i}\right\}\right)$ is the residual mapping to be learned. Our experiments show that the residual mapping is easier to optimize than the "plain net" counterpart, that simple stacked mapping.

\subsubsection{Attention Unit}
\begin{figure}[h]
\centering
\includegraphics[width=1.0\linewidth]{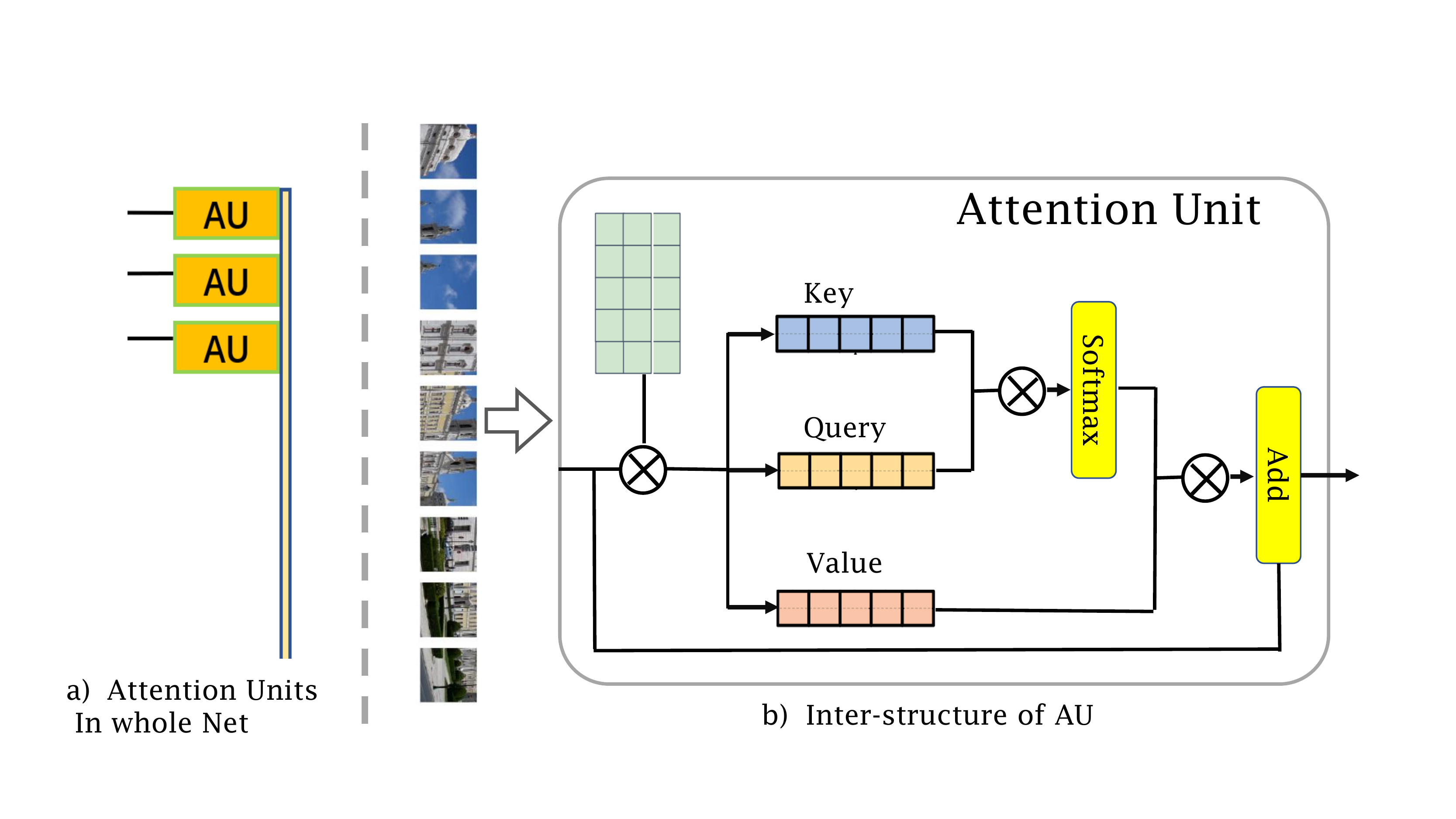}
\caption{Illustration of the downstream module Attention Unit $(a)$ and its Inter-Structure $(b)$ following by each of the convolutional location and label predictors.The implementation of AU follows similar with original Self-Attention mechanism.}
\label{AttentionUnit}
\end{figure}

Here, we place the attention unit (AU) after the fusion operation. The implementation Details is shown in Figure~\ref{AttentionUnit}.

Suppose that $\mathbf{x}^{\mathbf{s}} \in \mathbb{R}^{N^{s} \times C^{s}}$ is the output feature vectors at a specified scale $s \in\{1, \cdots, S\}$, in which $N$ and $C$ represent the scale of spatial locations and channels count in each feature, respectively. Firstly, we linearly calculate the input feature $x^{\mathbf{s}}$ with the training parameter $\mathbf{W}_{\mathbf{qkv}}^{\mathrm{s}}$ into three different feature spaces $key\left(x^{s}\right)$, $query\left(x^{s}\right)$, $value\left(x^{s}\right)$  by the matrix multiplication. We get the attention score by the matrix multiplication of $query\left(x^{s}\right) \text { and } key\left(x^{s}\right)$.
\begin{equation}
\mathbf{a}^{\mathbf{s}}=\mathbf{query}\left(\mathrm{x}^{\mathrm{s}}\right)^{\top} \mathrm{key}\left(\mathrm{x}^{\mathrm{s}}\right)
\end{equation}

Then the attention score matrix will be normalized by a softmax operation:
\begin{equation}
    \bar{a}_{i j}^{s}=\frac{\exp \left(a_{i j}^{s}\right)}{\sum_{j}^{N^{s}} \exp \left(a_{i j}^{s}\right)}, i, j=1,2, \cdots, N^{s}
\end{equation}

Finally, we calculate the attentive features (the Regions of Interest) by the matrix multiplication between $\mathrm{value}\left(\mathrm{x}^{\mathrm{s}}\right)$ and the attention weights $\overline{\mathbf{a}}^{\mathrm{s}}$. The weighted sums of individual features at each location is computed by using the following formula:
\begin{equation}
\mathbf{x}^{\mathbf{s}^{\prime}}=\mathbf{x}^{\mathbf{s}}+\left( \mathbf{value}\left(\mathbf{x}^{\mathbf{s}}\right)^{\top}\overline{\mathbf{a}}^{\mathbf{s}}\right)^{\top}
\end{equation}

The relevant parts of the feature map will be highlighted and the detection results will be refined through the Attention Unit (AU). Our comprehensive experiment indicates that the Attention Unit and the fusion mechanisms are complementary to each other.

\section{Experimental Result}

\subsection{Dataset Introduction}
MS COCO \cite{lin2015mscoco} and PASCAL VOC is the most popular open-source object detection datasets in this research field. Each image is annotated with bounding boxes and panoptic segmentation. We prefer to make some necessary combinations of these datasets to improve the model's prediction performance.

\subsubsection{Microsoft COCO}
COCO is a supervised dataset consisting of 1.7 million instances, with about 118k train$\&$value instances and up to 860k bounding boxes. Furthermore, there are about 7 object instances per image averagely and up to 63 instances in a single training sample. COCO has 53 stuff categories in addition to 80 object categories. Most of the state-of-the-art works and baselines are established on the challenging COCO object detection dataset. COCO can be downloaded through this download link: \href{https://cocodataset.org/\#home}{https://cocodataset.org/\#home}. 

\subsubsection{PASCAL VOC}
The PASCAL Visual Object Classes (VOC) challenge is relatively small benchmark dataset in visual object detection. VOC series have 20 object categories which combine of four main categories: vehicles, household, animals, person. There are total 27450 and 23080 trainval instances in VOC 2007 and VOC 2012, respectively. It is organised annually presented from 2005 while the popular parts of which are VOC 2007 and 2012. For the sake of simplicity, we will use appropriate abbreviations on VOC: 1) "07": VOC2007 trainval, 2)"07+12": the union set of VOC2007 and VOC2012 trainval, 3)"07+12+COCO": the union set of VOC2007, VOC2012 for training, then fine-tuning on COCO2014. All images and annotations are available at: \href{http://host.robots.ox.ac.uk/pascal/VOC/}{http://host.robots.ox.ac.uk/pascal/VOC/}.

\subsection{Implementation Details}
During our experiments, We choose to use many of the same hyper-parameter settings for CvT-ASSD as in original SSD. We convert the scale of the picture to a fixed size: 384$*$384 in order to ensure consistency of model input. Before training, we initialize the pretrained transformer backbone parameters by training on ImageNet-22k classification task and apply the Xavier-Uniform \cite{bingo_xavier} initialization method to other layers. In training step, we revise model by using Stochastic Gradient Descent (SGD) optimizer with initial rate $10^{-4}$ and a cosine learning rate decay scheduler. We also apply gradient clipping, with a maximal gradient norm of 0.05. Finally, at inference time we apply a final round of non-maximum suppression (NMS) with threshold 0.5 to filter our final detections. Our model`s pytorch implementation is released at github link \href{https://github.com/albert-jin/CvT-ASSD}{https://github.com/albert-jin/CvT-ASSD} for anyone doing experimental realization. Other relative implementation details are listed as follows:

\subsubsection{Data augmentation}
To make the model more robust to various input object sizes, shapes and contrasts, we use a more extensive sampling strategy, similar to SSD \cite{liuwei2016ssd}. Each training image is randomly sampled by one of the following options:

$-$ Use the original image as input.

$-$ Randomly crop with probability 0.5 to a rectanglar patch.

$-$ Sample a patch so that the minimum jaccard overlap: 0.1, 0.3, 0.5, 0.7, or 0.9 between objects.

$-$ Sample patches from per-image randomly.

$-$ Apply Contrast Enhancements to the entire original images.

Through above operations, we obtain diverse images for training. Our train-time scale data augmentation significantly improves the performance on small objects, indicating that the data augmentation trick is important for the final model accuracy.

\subsubsection{Transformer Internal Blocks}
Deep neural networks naturally integrate low/mid/high level features in an end-to-end multilayer model and the number of stacked layers (depth) can enrich the features levels.  We apply three stages to make up the CvT and design Backbone-Net CvT with different scales of parameters by varying the number of Transformer blocks of each stage and the hidden feature dimension used. Details of Internal Transformer structure is shown in Table~\ref{table_inter_cvt}.

\setlength{\tabcolsep}{1.0mm}
\begin{table}
\centering
\caption{This illustration shows the backbone of CvT-ASSD: Convolutional Transformer internal architecture. Inside our backbone, a different number of modules are stacked at each different stage.}
\label{table_inter_cvt}
\begin{tabular}{|l|c|c|}
\hline
\textbf{Stage. 1}  & \textbf{Embed}  &  \textbf{Blocks}\\ \hline
Details & 
7 × 7, 64, stride 4 & 
$\left[\begin{array}{c}
3 \times 3,64 \\
H_{1}=1, D_{1}=64 \\
R_{1}=4
\end{array}\right] \times 1$ \\ \hline
\textbf{Stage. 2}  & \textbf{Embed} & \textbf{Blocks}\\ \hline
Details & 
 3 × 3, 192, stride 2  &
$\left[\begin{array}{c}
3 \times 3,192 \\
H_{2}=3, D_{2}=192 \\
R_{2}=4
\end{array}\right] \times 4$ \\ \hline
\textbf{Stage. 3}  & \textbf{Embed} & \textbf{Blocks} \\ \hline
Details & 
3 × 3, 384, stride 2 &
$\left[\begin{array}{c}
3 \times 3,192 \\
H_{3}=6, D_{3}=384 \\
R_{3}=4
\end{array}\right] \times 16$ \\ \hline
\end{tabular}
\end{table}

\subsubsection{Bounding Boxes Settings}
We associate a set of default anchor boxes with each feature maps at the top of the network. These anchors actually are fixed multi-scale bounding boxes attached to each detection layers. We construct these default bounding boxes by defined box scales and aspect ratios. For each of 7 detection layers in our model, we successively apply aspect ratios with: $2, 2, [2, 3], [2, 3], [2, 3], 2, 2$ and box min/max sizes with: $[21,42], [42, 63], [63, 114], [114, 163], [163, 214], [214, 265]$, [265, 315], similar to SSD \cite{liuwei2016ssd}.

\subsection{Experimental Results}
For experimental comparison with baselines and SOTAs, we compare our model with several representative CNN-based approaches: VGG-SSD \cite{liuwei2016ssd}, Faster-RCNN \cite{ren2016fasterRCNN}, R-FCN \cite{dai2016rfcn} and Transformer-based models that have recently gained significant influence: ViT-FRCNN \cite{joshbeal2020vitfrcnn}, DETR \cite{FacebookAI2020detr}.

$-$ VGG-SSD (original) This framework is the first pure CNN-based single-shot (one-staged) object detector to perform comparably to state-of-the-art works on image object detection. The traditional approach provides the optimal trade-off among speed, accuracy and simplicity.

$-$ Faster-RCNN This work introduces a Region Proposal Network (RPN) on the top of Fast-RCNN \cite{girshick2014fastrcnn} that shares full-image convolutional features with the detection network, thus wins the 1st-place in ILSVRC and COCO 2015 competitions.

$-$ R-FCN (Region-based Fully Convolutional Network) A ResNet-based framework proposed by Microsoft, which adoptes the popular strategy that consists of two stages: region proposal and region classification. It inferences much quickly than the Faster-RCNN counterpart while achieves accuracy competitive than Faster-RCNN.

$-$ ViT-FRCNN A competitive solution which utilizes a transformer backbone on complex vision tasks such as object detection and segmentation. This work demonstrates the capability of Transformer-based models which pretrained with massive datasets can be fine-tuned to new relative tasks quickly.

$-$ DETR The DEtection TRansformer framework, proposed in 2020, consists of a transformer encoder-decoder architecture, and a set-based global loss that forces unique predictions via bipartite matching. It significantly outperforms competitive baseline like Faster-RCNN on the challenging COCO dataset.

For ablation study, Our CvT-ASSD can be divided into several variants with appropriate sub-module modifications. The corresponding model differences are listed as follows:

$-$ CvT-ASSD Our complete model which applied all features provided by this paper, including modules: CvT backbone, Residual Block, Attention Unit.

$-$ ViT-ASSD  We replace the transformer backbone with ViT backbone on our model for ablation study.

$-$ CvT-SSD  We remove the module Attention Unit from our complete model for ablation study.

$-$ CvT-ASSD$_{(noResidual)}$ We remove the module Residual Block from our complete model for ablation study.

\begin{table*}[t]
\caption{Detailed detection results on the PASCAL VOC2007 test set (4954 images). Blodface indicate scores better than other listed methods.Rows 1-4 shows baselines and SOTAs prediction performances. Rows 5-8 present our model variants prediction performances. }
\label{experiments_voc}
\centering
\begin{tabular}{l|l|llllllllllllllllllll} \hline
method & mAP & aeroplane & bike & bird & boat & bottle & bus & car & cat & chair & cow & table & dog & horse & mbike & person & plant & sheep & sofa & train & tv \\ \hline
Faster R-CNN & 76.4 & 79.8 & 80.7 & 76.2 & 68.3 & 55.9 & 85.1 & 85.3 & \bf{89.8} & 56.7 & \bf{87.8} & 69.4 & 88.3 & \bf{88.9} & 80.9 & 78.4 & 41.7 & 78.6 & \bf{79.8} & 85.3 & 72.0 \\
R-FCN & \bf{79.5} & 82.5 & 83.7 & \bf{80.3} & 69.0 & 69.0 & 69.0 & \bf{88.4} & 88.4 & 65.4 & 87.3 & 72.1 & 87.9 & 88.3 & 81.3 & 79.8 & 54.1 & \bf{79.6} & 78.8 & 87.1 & \bf{79.5} \\
VGG-SSD & 76.8 & 83.4 & \bf{84.7} & 78.4 & \bf{73.8} & 53.3 & 86.2 & 87.5 & 86.0 & 57.8 & 83.1 & 70.2 & 84.9 & 85.2 & 83.9 & 79.7 & 50.3 & 77.9 & 73.9 & 82.5 & 75.3 \\
DETR & 75.2 & 80.3 & 82.1 & 77.7 & 67.3 & \bf{57.2} & 79.3 & 85.1 & 83.7 & 64.5 & 84.3 & \bf{73.4} & 82.8 & 85.4 & 78.7 & 79.2 & \bf{53.3} & 76.7 & 72.1 & 84.7 & 74.5 \\ \hline
ViT-ASSD & 76.9 & 81.4 & 83.0 & 78.1 & 70.5 & 53.4 & 82.7 & 84.3 & 87.1 & 63.3 & 87.2 & 68.2 & 86.4 & 86.1 & 83.5 & 77.4 & 53.0 & 75.7 & 76.1 & 83.2 & 75.6 \\
CvT-SSD & 77.4 & 82.1 & 82.4 & 79.0 & 69.2 & 55.8 & 83.3 & 85.4 & 83.9 & 65.8 & 84.1 & 72.5 & 86.5 & 86.7 & 82.8 & 79.3 & 51.9 & 76.1 & 78.3 & 84.4 & 74.1 \\
CvT-ASSD$_{(noRsd)}$  & 77.6 & 82.8 & 84.0 & 79.3 & 71.3 & 55.7 & 83.6 & 85.8 & 84.1 & 66.2 & 84.6 & 72.9 & 87.2 & 87.3 & 82.3 & 80.1 & 52.1 & 74.8 & 78.7 & 83.6 & 74.7 \\
CvT-ASSD & 78.5 & \bf{83.7} & 84.2 & 79.8 & 71.7 & 56.3 & \bf{84.0} & 86.4 & 85.3 & \bf{67.4} & 86.7 & 73.3 & \bf{87.4} & 87.9 & \bf{84.6} & \bf{80.3} & 52.3 & 75.2 & 79.4 & \bf{85.5} & 77.6 \\ \hline
\end{tabular}
\end{table*}

\subsubsection{Object Detection on COCO}
\begin{table}[h]
\renewcommand{\arraystretch}{1.2}
\caption{Comparisons between our model variants and other famous models on MS COCO dataset. The COCO-style AP is evaluated $@ IoU \in [0.5, 0.95]$.  Blodface indicate scores better than other listed methods.}
\label{experiments_coco}
\centering
\begin{tabular}{c||c|c|c|c|c}  \hline
method & mAP & AP$_{small}$ & AP$_{medium}$ & AP$_{large}$ & \#params \\ \hline
Faster-RCNN & 27.2 & 6.6 & 28.6 & 45.0 & 60M \\ \hline
R-FCN & 27.6 & 8.9 & 30.5 & 42.0 & 55.9M \\ \hline
VGG-SSD(original) & 27.9 & 8.3 & 30.3 & 45.1 & 52.0M \\ \hline
ViT-FRCNN & 37.8 & 17.8 & 41.4 & 57.3 & 46.2M \\ \hline
DETR & \bf{42.0} & 20.5 & 45.8 & \bf{61.1} & 37.4M\\ \hline \cline{1-6}
ViT-ASSD & 35.2 & 17.5 & 42.8  & 47.9 & 44.0M\\ \hline
CvT-SSD & 38.2 & 19.4 & 45.2 &  52.8 & \bf{29.6M} \\ \hline
CvT-ASSD$_{(noResidual)}$ & 38.9 & 20.1 & 45.5 & 55.3 & 30.1M \\ \hline
CvT-ASSD & 41.3 & \bf{21.2} & \bf{46.3} & 56.3 & 32.7M \\  \hline
\end{tabular}
\end{table}

In this section, We conduct our experiments on Microsoft COCO 2017 which contains about 12w images. We split it into three parts: 10w train set, 1w validation set and 1w test-dev set. We employ SGD optimizer for 40 epochs using a cosine decay learning rate scheduler when training from scratch. Our model is trained on 2 GPUs with 8 images per GPU for 400000 iterations. We make comparisons between our model and previous baseline models to prove the effectiveness of our model. Table~\ref{experiments_coco} reports the comparison of our best results with those of previous state-of-the-art frameworks on MS COCO. As depicted in this table, our model outperforms most other baselines such as ResNet-based Faster-RCNN by improving AP(+12.9), AP$_s$(+14.6), AP$_m$(+17.7), AP$_l$(+11.3) on MS COCO dataset and requires fewer parameters \textbf{32.7M} than the Faster-RCNN counterpart \textbf{60.0M}. Meanwhile, CvT-ASSD achieve comparable results to ViT-FRCNN and DETR, while having fewer parameters. Perhaps more interestingly, our model obtains performance of \textbf{41.3mAP} in MS COCO test dataset, which performs not so well compared with SOTA method, the DETR of \textbf{42.0mAP}. This phenomenon can be attributed to the shallower layer of our transformer backbone than DETR backbone ViT. In the future, we will try to deepen our CvT backbone depth in order to achieve more accurate image understanding.

As shown in Table~\ref{experiments_coco}, our complete model CvT-ASSD loses 0.7mAP to the state-of-the-art method DETR but achieves greater performance when detecting smaller object. While CvT-ASSD may not achieve state-of-the-art results on COCO, we believe this signifies the possibility and superiority of introducing convolutions into transformer and applying transformer to SSD approach as its backbone.

Furthermore, experimental result gaps between CvT-ASSD and its variants indicate that our new introduced components (CvT, RB and AU) all significantly contribute to the final object detection performance.

\subsubsection{Object Detection on VOC}
\begin{figure*}[h]
\centering
\includegraphics[width=1.0\linewidth]{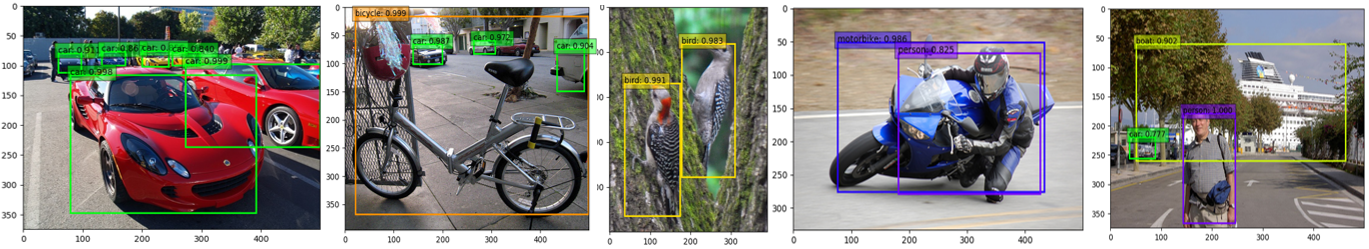}
\caption{Visualizing CvT-ASSD prediction results for objects belong to different categories (images from PASCAL COCO test set). Objects in each image are shown with their predicted boxes and corresponding category label clearly.}
\label{visualization}
\end{figure*}

In this section, we carry out our experiment for training models on PASCAL VOC2007 trainval and VOC2012 trainval dataset (VOC07+12), and validate these models on VOC2007 test dataset (\textbf{4954} images). Especially, we compare against Faster-RCN, R-FCN, original SSD, DETR by the VOC-style average precision ($\%$) metric. From Table~\ref{experiments_voc}, It is obvious that CvT-ASSD performs well when compared with several baselines and it is very robust to different object aspect ratios because we use default boxes of various aspect ratios per feature map location. In particular, we are excited about CvT-ASSD`s capability  to transform representations training on huge scale classification datasets ImageNet-22k to improved performance on object detection tasks.

We can clearly see that our model has better performance on smaller objects than bigger objects. This is not surprising because we apply shallow-level features with more bounding boxes for prediction instead of high-level features. What surprises us is that we can find our model achieves state-of-the-art particular performances in several categorises: achieving \textbf{83.7mAP}  relative to aeroplane, achieving \textbf{67.4mAP} relative to chair and so on.

In ablation part, as seen in Table~\ref{experiments_voc} Row5-8, the complete model CvT-ASSD significantly outperforms other ablation models. For example, It achieves high performance up to \textbf{78.5mAP}, increasing \textbf{1.1mAP} compared to the CvT-ASSD which removes the Attention Units (AUs) and increasing \textbf{2.6mAP} compared to the ViT-ASSD which replaces traditional naive transformer ViT \cite{dosovitskiy2021ViT} with our CvT. The performance improvement confirms that our vision-transformer variant CvT, Residual Block (RB) and Attentive Unit (AU) can both improve the average precision performance for object detection task, similar to validation on MS COCO. Furthermore, It proves that residual operation and self-attention mechanism can help increase network convergence ability and final model performance.

Meanwhile, we display our CvT-ASSD prediction results visualization in Figure~\ref{visualization}. The colored quadrilaterals are the object positioning results and the corresponding text labels in the upper corner of bounding boxes are the object category prediction confidences.

\section{Conclusion}
In this work, we present CvT-ASSD, a competitive, simple but efficient object detection approach which apply Convolutional vision Transformer as detector backbone, residual module and self-attention mechanism to original SSD detection structure. Comprehensive experiments show that our approach achieves comparable performances to a few baselines like Faster-RCNN on the PASCAL VOC and MS COCO datasets. With a fewer parameters and competitive accuracy compared with baselines even SOTAs, we believe our proposed model can provide a useful real-time object detection component for large artificial intelligence system applications. We hope that our approach will inspire the exploration of convolutional transformer-based models for more complex visual tasks in the future.


\section*{Acknowledgment}
% This work was started as an anonymous scientific research project at Shanghai University. We would like to thank Xiangfeng Luo for helpful discussions and comments. We also thank NVIDIA for the sufficient computing power supported from two GTX-1080TI GPUs.
The research reported in this paper was supported in the Outstanding Academic Leader Project of Shanghai under the grant No.20XD1401700 and part by the National Natural Science Foundation of China under the grant 91746203 and the Ministry of Industry and Information Technology project of the Intelligent Ship Situation Awareness System under the grant No.MC-201920-X01. We would like to thank Xiangfeng Luo for helpful discussions and comments.



%\bibliographystyle{IEEEtran}

\end{document}